%% file: main.tex
\documentclass[final,1p,times]{elsarticle}
\usepackage{amssymb}
\usepackage{amsmath}
\journal{Journal of Computer Vision and Image Understanding}

\usepackage[table,xcdraw]{xcolor} % colour support
\usepackage{colortbl} % table colour

\usepackage{tabularx}
\usepackage{booktabs} % for better table lines

\usepackage[colorlinks=true, citecolor=blue, linkcolor=black, urlcolor=blue]{hyperref}
\usepackage{caption}
\usepackage{subcaption}

\begin{document}

\begin{frontmatter}



\title{Gloss-Free Sign Language Translation: An Unbiased Evaluation of Progress in the Field}


\author{Ozge Mercanoglu Sincan} %% Author name
\author{Jian He Low}
\author{Sobhan Asasi}
\author{Richard Bowden}
\affiliation{organization={Centre for Vision, Speech and Signal Processing (CVSSP), University of Surrey},%Department and Organization
        %    addressline={}, 
            city={Guildford},
            postcode={GU2 7XH}, 
            state={Surrey},
            country={UK}} 

%% Abstract
\begin{abstract}
Sign Language Translation (SLT) aims to automatically convert visual sign language videos into spoken language text and vice versa. 
 While recent years have seen rapid progress, the true sources of performance improvements often remain unclear. Do reported performance gains come from methodological novelty, or from the choice of a different backbone, training optimizations, hyperparameter tuning, or even differences in the calculation of evaluation metrics? This paper presents a comprehensive study of recent gloss-free SLT models by re-implementing key contributions in a unified codebase. We ensure fair comparison by standardizing preprocessing, video encoders, and training setups across all methods. Our analysis shows that many of the performance gains reported in the literature often diminish when models are evaluated under consistent conditions, suggesting that implementation details and evaluation setups play a significant role in determining results. We make the codebase publicly available here \footnote{https://github.com/ozgemercanoglu/sltbaselines} to support transparency and reproducibility in SLT research. 
\end{abstract}

%Graphical abstract

%Research highlights
%\item Unified benchmark of recent gloss-free sign language translation models
%\item Examining key innovations and an unbiased evaluation of progress in the field
%\item Open-source codebase for reproducibility

%Keywords

\end{frontmatter}



%%%%%%%%%%%%%%%%%%%%%%%%%%%%%%%%%%%%%%%%%%%%%%%%%%%%%%%%%%%%%%%%%%%%%%%%%%%%%%%5
\input{sections/1_intro}

\input{sections/2_relatedwork}

\input{sections/3_method}

\input{sections/4_experiments}
\input{sections/5_conclusion}
\section*{Acknowledgments}
\label{sec:Acknowledgemnts}
This work was supported by the SNSF project ‘SMILE II’ (CRSII5 193686), the Innosuisse IICT Flagship (PFFS-21-47), EPSRC grant APP24554 (SignGPT-EP/Z535370/1), and through funding from Google.org via the AI for  Global Goals scheme. This work reflects only the author’s views and the funders are not responsible for any use that may be made of the information it contains.

%% Displayed equations can be tagged using various environments. 
%% Single line equations can be tagged using the equation environment.
%\begin{equation}
%f(x) = (x+a)(x+b)
%\end{equation}

%% Refer following link for more details.
%% https://en.wikibooks.org/wiki/LaTeX/Mathematics
%% https://en.wikibooks.org/wiki/LaTeX/Advanced_Mathematics

%% Use a table environment to create tables.
%% Refer following link for more details.
%% https://en.wikibooks.org/wiki/LaTeX/Tables


\bibliographystyle{elsarticle-num.bst}
\bibliography{refs}
\end{document}

%% file: sections/1_intro.tex
\begin{figure*}[htbp]
        \centering
        \includegraphics[width=\linewidth]{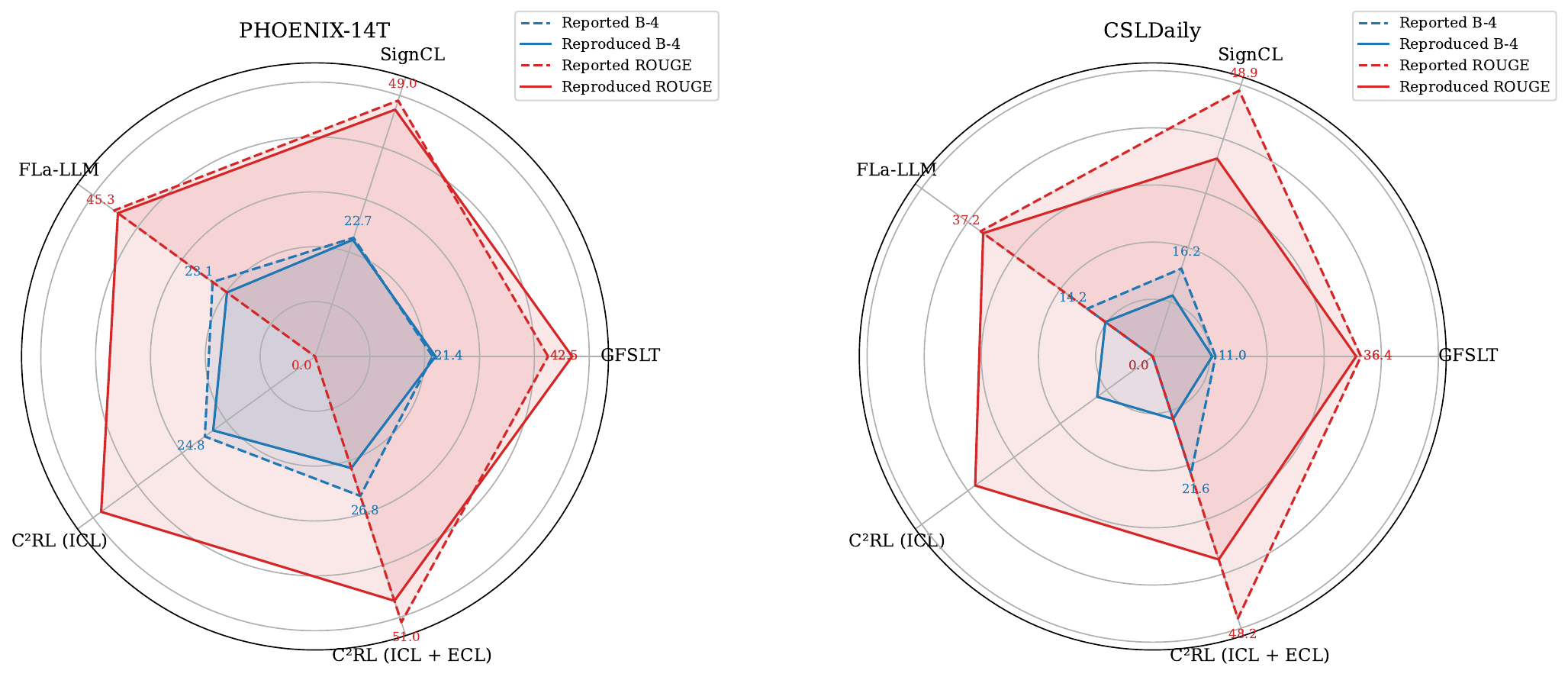}
    \caption{Reported vs reproduced results.}
    \label{fig:scores}
\end{figure*}

\section{Introduction}
\label{sec:intro}

Sign languages are visual languages that are mainly used by the deaf community. Like spoken languages, they vary between countries, and each sign language has its own lexicon and grammatical structures \cite{sutton1999linguistics}. However, the structural and grammatical differences are even greater between sign and spoken languages. Sign languages involve multiple simultaneous channels of communication, such as hand gestures, facial expressions, mouthings and mouth gestures\footnote{In sign language linguistics, mouthing refers to the silent articulation of spoken language words alongside signs, often reflecting the local spoken language, while mouth gestures are non-verbal movements of the mouth that are part of the sign itself and convey grammatical or affective information.} and body posture. All of which occur in parallel during the signing process. In contrast, spoken and written languages are typically linear, where words and phrases are articulated one after the other. Moreover, the word order of spoken languages usually follows syntactic rules, such as subject-verb-object (SVO), whereas in sign languages, the order of signs often differs, and the use of space (both topological and syntactical) can result in profound variations in how something can be expressed in sign language.

Designing intelligent systems that can bridge the communication gap between deaf and hearing individuals is a critical step toward creating inclusive and accessible technologies. Sign Language Translation (SLT) plays a central role in this context, aiming to convert sign language into spoken language and vice versa. This task generally involves two main directions: (1) sign-to-text, where visual input (e.g., video of a signer) is translated into a spoken language sentence \cite{camgoz2018neural}, and (2) text-to-sign, where an input spoken sentence is converted into sign language using avatars \cite{cox2002tessa,mcdonald2016automated}, or photo-realistic video synthesis \cite{saunders2022signing}. In this paper, we focus on the sign-to-text direction.

Sign-to-text translation remains challenging for several reasons: (i) the structural and grammatical differences between sign and spoken languages, (ii) the limited availability of large-scale, annotated datasets, and (iii) the visual complexity and variation in signing styles. Sign languages exhibit high intra- and inter- signer variability in terms of speed, articulation, and expression. Successful models must therefore learn both fine-grained visual representations (handshape trajectories, facial action units, body posture) and high-level semantic abstractions that generalize across signers and linguistic contexts.

Previous works have relied on intermediate gloss\footnote{Gloss is a written representation of a sign using words from spoken language.} annotations, where glosses serve as a linguistic bridge between sign videos and spoken-language sentences \cite{camgoz2018neural, zhou2021improving, chen2022two, yin2021simulslt, zhou2021spatial}. However, data annotation is expensive, and extreme care needs to be taken to ensure consistency in gloss annotation between annotators. It is labor-intensive and takes on average one hour per 90 seconds of video \cite{duarte2021how2sign}. Therefore, recent research has shifted towards gloss-free approaches, eliminating the need for gloss annotation and learning a direct mapping from sign language videos to text. These gloss-free pipelines typically integrate advanced video encoders with sequence modeling and attention mechanisms to capture the full richness of sign language \cite{zhou2023gloss, yin2023gloss, chen2024c, chen2024factorized, ye2024improving, kim2024leveraging, sincan2023context, li2025unisign}. One of the most influential models in this direction is GFSLT-VLP~\cite{zhou2023gloss}, which introduces a novel visual-language pretraining strategy to align visual and textual representations in a joint semantic space. Its public implementation has served as a foundation for many follow-up studies \cite{ye2024improving, chen2024factorized, chen2024c}.

As in many machine learning domains, researchers are often competing to maximize benchmark performance. However, behind each reported number lies a multitude of design choices, such as selection of backbone architectures, loss functions, and regularization techniques. All these decisions significantly influence the performance of SLT models. However, the papers often emphasize novel theoretical contributions over comprehensive reporting of implementation details and hyperparameter settings, making it difficult to disentangle the true impact of core design decisions.

Our work contributes to this direction by analyzing and fairly comparing recent gloss-free sign language translation methods--GFSLT-VLP \cite{zhou2023gloss}, SignCL \cite{ye2024improving}, Sign2GPT \cite{wong2024signgpt}, FLa-LLM \cite{chen2024factorized}, and C2RL \cite{chen2024c}-- within a unified, modular codebase, most of which extend or refine the GFSLT-VLP framework. By standardizing data preprocessing, video encoders, and optimization schedules, we ensure all approaches are trained and evaluated under the same conditions. As a result, our study highlights the impact of key design choices on model performance. Figure \ref{fig:scores} compares published performance results with those obtained in our re-implementation on the Phoenix-2014T \cite{camgoz2018neural} and CSL-daily \cite{zhou2021improving} benchmarks, offering insights into the reproducibility and robustness of current SLT methods. Furthermore, our objective is to make this codebase and data processing framework available to the wider scientific community, so we can collectively progress the field.

The contributions of this paper can be summarized as:
\begin{itemize}
    \item We provide a modular codebase containing five state-of-the-art gloss-free SLT models, standardizing core components such as data preprocessing pipelines, video encoders, and training schedules.
    \item We examine each model’s key innovations --visual-language pretraining, contrastive learning on adjacent frames, pseudo-gloss pretraining, lightweight translation in pretraining, and hybrid CLIP-translation in pretraining-- to measure their isolated impact on translation quality.
    \item We evaluate the methods on two widely used corpora, Phoenix-2014T and CSL-daily, ensuring identical evaluation protocols.

\end{itemize}

%% file: sections/2_relatedwork.tex
%------------------------------------------------------------------------
\section{Related Work}

Sign language research has progressed from basic sign recognition to continuous recognition and to translation, with each stage introducing more complexity and linguistic nuance. Sign Language Recognition (SLR) focuses on identifying individual signs from video frames, typically in an isolated setting \cite{joze2018ms, li2020word, sincan2020autsl, jiang2021skeleton, hu2021hand, zuo2023natural}. Building on isolated recognition, Continuous Sign Language Recognition (CSLR) seeks to segment and transcribe unsegmented signing streams into sequences of gloss tokens \cite{min2021visual, zhou2021improving, wei2023improving, hu2023continuous}. Sign Language Translation (SLT), sign-to-text, aims to generate grammatically coherent sentences in a target spoken language given sign language videos.

In this section, we discuss relevant works on gloss-based and gloss-free sign language translation, as well as sign language datasets.

\subsection{Gloss-based Sign Language Translation}

Camgoz et al. \cite{camgoz2018neural} first formulated SLT as a sequence-to-sequence task using a CNN encoder and an RNN decoder, and released the first SLT dataset, Phoenix-2014T, which contained gloss sequences. Due to the sequential alignment between gloss annotations and the temporal structure of sign language videos, early SLT research heavily relied on gloss-level supervision to simplify the translation task. For instance, SLRT \cite{camgoz2020sign} introduced a transformer encoder-decoder framework, integrating CTC loss \cite{graves2006connectionist} to soft-match sign representations and gloss sequences. Many subsequent works have leveraged gloss annotations to improve SLT performance, either by incorporating them into end-to-end transformer-based models or by decomposing the task into two sequential modules: a sign-to-gloss recognition stage followed by gloss-to-text translation \cite{zhou2021improving, zhou2021spatial, chen2022two, chen2022simple, zhang2023sltunet, yin2021simulslt}. Some studies proposed modular pipelines that segment isolated signs using independently trained classifiers before feeding predictions into a translation module \cite{sincan2024using, zuo2024towards}. On the other hand, several studies concentrate specifically on the gloss-to-text sub-task, aiming to enhance spoken sentence generation from gloss input by leveraging visual and contextual information, pre-trained language models, or non-autoregressive modeling \cite{jing2024vk, fayyazsanavi2024gloss2text, zhou2025non}.

In these approaches, glosses act as an intermediate representation, reducing the complexity of direct sign-to-text translation. However, they require large-scale gloss-annotated datasets, which are expensive and time-consuming to create. In addition, the two-stage approach is also prone to error propagation, where inaccuracies in gloss prediction can adversely affect the downstream translation. The reliance on fixed-vocabulary gloss classifiers may also constrain the model’s capacity to generalize to unseen or out-of-vocabulary signs.

\subsection{Gloss-free Sign Language Translation}
More recently, research has shifted towards gloss-free SLT that aims to directly translate sign language videos into spoken language without relying on gloss annotations \cite{gong2024llms}. Initial efforts trained single-stage sequence-to-sequence models with enhanced visual encoders—e.g., 3D‑CNNs or graph‑based pose features—and attention/transformer architectures \cite{yin2023gloss, lin2023gloss, sincan2023context}. 

Recent advances have further improved gloss-free SLT through the use of pretraining strategies. GFSLT-VLP \cite{zhou2023gloss} was one of the pioneering studies, introducing a CLIP-style  \cite{radford2021learning} contrastive learning strategy to learn alignments between sign videos and spoken language sentences. Due to its strong performance and publicly available implementation, it has become a widely adopted baseline, serving as the foundation for several models. Subsequent methods have explored various strategies, including contrastive learning between adjacent frames \cite{ye2024improving}, enriching visual representations with additional modalities \cite{kim2024leveraging, hwang2024efficient}, leveraging pseudo-glosses \cite{wong2024signgpt}, adopting generative pretraining paradigms \cite{chen2024factorized}, combining visual-language alignment with generative objectives in a multi-task setup \cite{chen2024c}, and converting signs into discrete representations \cite{gong2024llms}. Including the aforementioned methods, recent approaches also integrate large language models (LLMs) or generative pretrained transformers (GPTs) to improve the quality of the generated translations \cite{liang2024llava, jang2025lost}. 
Recent efforts have also explored leveraging large-scale external sign language datasets to enhance representation learning \cite{li2025unisign, rust2024towards}. These approaches ultimately share a common objective, which is to improve translation performance.

In this work, we focus on a comparative analysis of five recent methods \cite{zhou2023gloss, ye2024improving, wong2024signgpt, chen2024factorized, chen2024c}, which adopt different pretraining approaches, most of which extend or refine the GFSLT-VLP. This selection helps comparison to be fair and consistent. By evaluating these models under a consistent setting, we aim to better understand how various training strategies influence gloss-free SLT performance.

% -----------------------------------------

\subsection{Datasets}
\begin{figure}[htbp]
    \centering
    \includegraphics[width=0.9\linewidth]{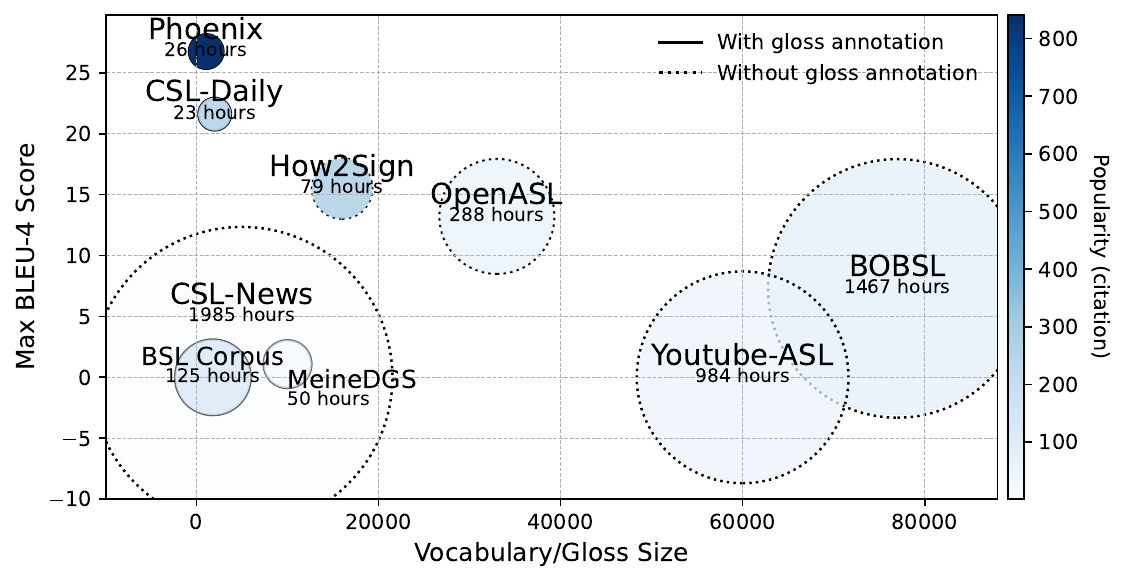}
    \caption{Comparison of sign language translation datasets in terms of gloss/vocabulary size, BLEU-4 score, dataset duration, and citation-based popularity. The x-axis represents the gloss size for datasets with gloss annotations (shown with solid borders), or vocabulary size from spoken language sentences for datasets without gloss annotations (shown with dashed borders). The y-axis shows the maximum BLEU-4 translation performance as reported in the literature. Circle sizes reflect the total number of hours. Color opacity encodes dataset popularity based on citation counts, with darker circles indicating more widely used datasets.}
    \label{fig:datasets}
\end{figure}

Although numerous datasets have been developed for isolated and continuous sign language recognition \cite{joze2018ms, li2020word, sincan2020autsl, forster2012rwth, desai2023asl}, datasets specifically designed for sign language translation have only recently seen notable growth. Although some linguistic projects were initiated in 2008, such as DGS Corpus and BSL Corpus, releases of data and annotations became available in recent years \cite{Schembri2017British, konrad2020meine}. An overview of SLT datasets is provided in Table~\ref{tab:dataset}, including their source, scale, annotation format, popularity, and highest performance in the literature. Additionally, a visual comparison of these datasets is presented in Fig. \ref{fig:datasets}, providing a clear understanding of their characteristics.

\begin{table*}
	\centering
         \resizebox{1\linewidth}{!}{
        	\begin{tabular}{ lccccccccc}
        		\hline   
        		\textbf{Dataset} & \textbf{Year}  & \textbf{Source}  & \textbf{Domain} &  \textbf{Signers} & \textbf{Hours} & \textbf{Text Vocab.} & \textbf{Glosses}  & \textbf{Max BLEU4} & \textbf{Popularity†}  \\ \hline
                BSL Corpus \cite{schembri2013building, Schembri2017British} & 2008-present* & Lab & Conversational & 249 & 125 & - & $\sim$1800 & - & 187 \\
                Phoenix-2014T \cite{camgoz2018neural} & 2018  & TV & Weather & 9 & 11 & 3K &1066 &  26.75 \cite{chen2024c} & 841 \\ 
                MeineDGS \cite{konrad2020meine} & 2008-20* & Lab & Conversational & 330 & 50 & 18K & $\sim$10000 & 1.08 \cite{sincan2024using} & 16 \\
                CSL-Daily \cite{zhou2021improving} & 2021 & Lab & Daily & 10 & 23 & 2K & 2000 &  21.61 \cite{chen2024c} & 256 \\
                How2Sign \cite{duarte2021how2sign} & 2021  & Lab & Instructional & 11 & 79 & 16K & N/A & 15.5 \cite{rust2024towards} & 262 \\
                BOBSL \cite{albanie2021bbc} & 2021 & TV & Diverse & 39 & 1467 & 77K & N/A & 7.3 \cite{jang2025lost} & 82 \\
                OpenASL \cite{shi2022open} &  2022 & Web & Open & 220 & 288 & 33K & N/A & 13.21 \cite{chen2024c} & 62 \\     
                Youtube-ASL \cite{uthus2023youtube} & 2023 & Web & Open & 2519 & 984 & 60K & N/A & - & 52 \\
                YouTube-SL-25 \cite{tanzer2024youtube} & 2024 & Web & Open & 3000 & 3207 & - & N/A & - & 6 \\
                CSL-News \cite{li2025unisign} & 2025 & TV & Diverse & - & 1985 & 5K & N/A & - & 1 \\
                
            \hline       
        	\end{tabular}
         }
 	\caption{Summary statistics of sign language datasets. N/A demonstrates that gloss annotations are not available for the dataset, `-' indicates that the value was not reported. *The BSL Corpus, initiated in 2008, and its data annotation is still in progress. MeineDGS was initiated in 2008. †Popularity is the number of citations for the associated publication(s) as of May 13, 2025.}
     \label{tab:dataset} 
	\end{table*}

\textbf{BSL Corpus} is a collection of approximately 125 hours of British Sign Language data signed by 249 signers from 8 regions around the UK. It includes narratives, lexical elicitation, interviews, and conversational data. Lexical-level annotations and some of the English translations are available, and more annotations are being made available periodically. \textbf{MeineDGS} is also a linguistic dataset on German Sign Language, containing 50 hours of video from 330 signers. Similar to the BSL Corpus, it was designed with a focus on regional and linguistic diversity. The data format is diverse, ranging from story-retelling to free-flowing conversations. The dataset provides both gloss and sentence-level annotations.

\textbf{Phoenix-2014T} \cite{camgoz2018neural} provides gloss annotations and spoken language sentences for German Sign Language (DGS) videos. Since its release, it has become a benchmark and played a central role in shaping SLT research. However, it is limited to a narrow domain of weather forecasts and a limited vocabulary. It has a vocabulary size of 1066 signs, 2887 German words, and 7K sentences in the training set. To address these limitations, subsequent datasets aimed to expand both linguistic and topical diversity. \textbf{CSL-Daily} \cite{zhou2021improving} is designed for daily conversations in Chinese sign language, covering a diverse range of topics, such as family life, medical care, bank service, and shopping etc.  The dataset includes 23 hours of video recorded in a laboratory environment with 10 deaf signers. It has a vocabulary size of 2000 signs, 2343 Chinese words, and 18K sentences in the training set.

\textbf{How2Sign} \cite{duarte2021how2sign} expands the scope of SLT resources by providing multimodal data, including RGB, depth, speech, pose estimation, and multiview recordings for American Sign Language (ASL). Unlike earlier resources, How2Sign does not provide gloss annotations but instead contains English transcripts aligned with sign videos. The dataset consists of 80 hours of instructional videos in ASL in 10 categories, such as foods, hobbies, education, etc. The dataset is recorded in a laboratory environment with 11 professional hearing or deaf individuals.

In the meantime, efforts have been made to scale data collection and bring more real-world diversity into SLT \cite{albanie2021bbc, li2025unisign}. \textbf{BOBSL} \cite{albanie2021bbc}, a large-scale British Sign Language dataset, is curated from interpreted BBC broadcast footage, totalling 1,467 hours with a wide range of topics. However, the dataset provides only audio-aligned English subtitles - manual alignment is performed for only the test set (31 hours). Similarly, the CSL-News \cite{li2025unisign}, Chinese Sign Language (CSL) dataset, contains 1,985 hours of video paired with textual annotations from news content. \textbf{OpenASL} \cite{shi2022open} and \textbf{Youtube-ASL} \cite{uthus2023youtube} are collected from online platforms to provide additional large-scale training data, though they include more noise. Youtube-ASL does not provide train, validation, and test splits; so, it serves as external data to increase sign language translation performance on other datasets like How2Sign.

Recently, \textbf{YouTube-SL-25} \cite{tanzer2024youtube} has focused on building multilingual sign language datasets to support cross-lingual SLT models. 

In this work, Phoenix-2014T and CSL-Daily datasets were chosen for their widespread use, and also due to their manageable size, which makes them suitable for reproducible and controlled experiments.

%% file: sections/3_method.tex
%------------------------------------------------------------------------
\section{Overview of Methods Evaluated}

%This section begins by formulating the sign language translation (SLT) problem, followed by an overview of the implemented SLT approaches. 

Given an input video $V = (x_1, x_2, ..., x_T)$ with $T$ frames, a sign language translation aims to translate the sign video into a spoken language sequence $S = (w_1, w_2, ..., w_U)$ with $U$ words. 

In recent years, one of the most significant advancements has been the introduction of visual-language pretraining techniques \cite{zhou2023gloss}. The availability of public implementation has further accelerated progress by enabling reproducibility and extensions, especially on small-scale datasets such as Phoenix-2014T \cite{camgoz2018neural}. 

Given the influence of GFSLT-VLP \cite{zhou2023gloss}, we examine state-of-the-art methods that extend or refine this approach, including SignCL \cite{ye2024improving}, FLa-LLM \cite{chen2024factorized}, C\textsuperscript{2}RL \cite{chen2024c}. These methods build upon a shared architectural pipeline while introducing distinct training objectives or strategies. Additionally, we include Sign2GPT \cite{wong2024signgpt}, a recent approach that incorporates pseudo-gloss pretraining strategy. While it follows a distinct network design, incorporating pseudo-gloss training allows us to explore the impact of intermediate representations.

\begin{figure*}[htbp]
    \centering
    \includegraphics[width=\linewidth]{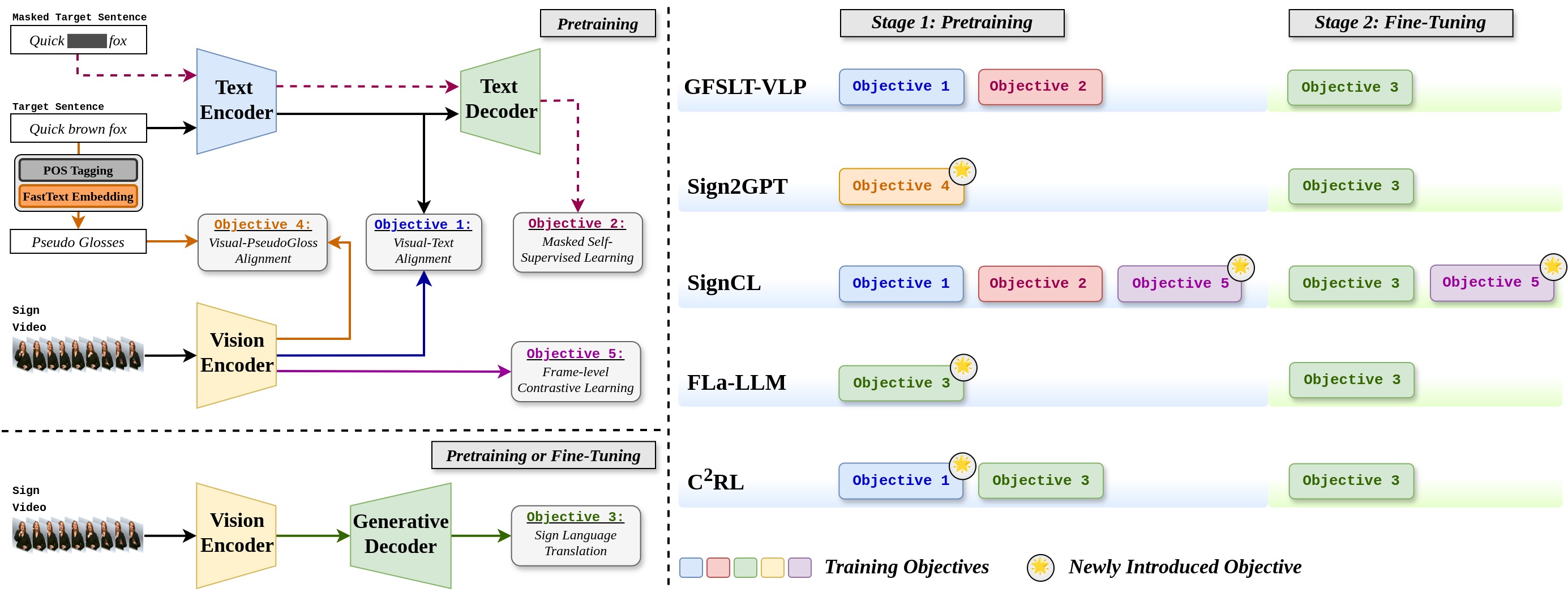}        
    \caption{Architectural overview and training objectives of the compared sign language translation methods. The right side summarizes which objectives are used by each method.}
  \label{fig:methods}
\end{figure*}

Figure~\ref{fig:methods} illustrates the common architectural pipeline across the models and highlights differences in their training objectives. Although the common structure is similar, methods differ in architectural choices and training setups. These differences are detailed in Table~\ref{tab:parameters_sota}.
Here, we summarize each method by highlighting their architectural designs, and training objectives. 

\begin{table*}
	\centering
         \resizebox{1\linewidth}{!}{
        \begin{tabular}{ lccccc}
            \hline   
            & \textbf{GFSLT-VLP} & \textbf{Sign2GPT} & \textbf{SignCL} &  \textbf{FLa-LLM} & \textbf{C\textsuperscript{2}RL}  \\ \hline
            Key idea & Visual-language pretraining  & Pseudo-gloss pretraining & Contrastive learning on frames & Light-T in pretraining &  Combine CiCO and Light-T  \\
            Code availability & Yes & Yes & Partially & No & No \\
             \rowcolor[HTML]{E2E2E2}
            \multicolumn{6}{l}{Pretraining stage}  \\ 
            number of epochs & 80 & 100 & 80 \textsuperscript{*}  & 100 & 200 \\ 
            visual backbone & ResNet18† & DinoV2 & Resnet18† & ResNet18† & ResNet18†  \\
            \quad -TempBlock & Conv1D-BN-ReLU-MaxPool & -- & Conv1D-BN-ReLU-MaxPool & Conv1D-BN-ReLU & Conv1D-BN-ReLU \\
            \quad -Temporal Encoder & 1024 hid, 4096 FF & -- & 1024 hid, 4096 FF\textsuperscript{*}  & 512 hid, 2048 FF & 512 hid, 2048 FF \\
            batch size & 8 & 8 & 128 & 16 & 64 \\ 
            optimizer & SGD & adam & SGD\textsuperscript{*}  & SGD & SGD \\
            scheduler & cosine & one-cycle cosine & cosine\textsuperscript{*}  & cosine & cosine \\
            learning rate & $10^{-2}$ & $3\times10^{-4}$ & $10^{-2}$ \textsuperscript{*}  & $10^{-2}$ & $10^{-2}$  \\ 
             label-smoothing & - & - & - & 0.2 & 0.2 \\
             dropout & 0.1 & & 0.1\textsuperscript{*}  & & 0.1 \\
            \hline
             \rowcolor[HTML]{E2E2E2}
            \multicolumn{6}{l}{Translation stage}  \\ 
            number of epochs & 200 & 100 & * & 75 & 80 \\
            LLM backbone & 3layers-Mbart & XGLM & 3layers-Mbart  & 12layers-Mbart & 12layers-Mbart  \\
            batch size & 8 & 8 & 128 & 16 & 128 \\ 
            optimizer & SGD & adam & SGD\textsuperscript{*}  & adam & adam \\
            scheduler & cosine & one-cycle cosine &  cosine\textsuperscript{*} & cosine & cosine \\
            learning rate & $10^{-2}$  & $3\times10^{-4}$  &  $10^{-2}$ \textsuperscript{*}  & $10^{-5}$, $10^{-3}$ & $10^{-5}$, $10^{-3}$  \\ 
            label-smoothing & 0.2 & 0.1 & 0.2\textsuperscript{*}  & 0.2 & 0.2\\ 
            dropout & & & & & 0.3 \\

            \hline
            \rowcolor[HTML]{E2E2E2}
            \multicolumn{6}{l}{Data processing}  \\ 
            frame sampling strategy & random & subsampling 25\% & random\textsuperscript{*}  & subsampling 50\%  & subsampling 50\%  \\
            max number of frames & 300 &  & 300\textsuperscript{*}  \\
            \hline     
            \rowcolor[HTML]{E2E2E2}
            \multicolumn{6}{l}{Scores}  \\ 
            BLEU library & nlgeval & sacrableu & nlgeval\textsuperscript{*}  &  &  \\
            BLEU-1 & 43.71 & 49.54 & 49.76 & 46.29 & 52.81 \\
            BLEU-4 & 21.44 & 22.52 & 22.74 & 23.09 & 26.75\\
            ROUGE & 42.49 & 48.94 & 49.04 & 45.27 & 50.96\\
            \hline     
            
        \end{tabular}
         }
 	\caption{Architectural design and hyperparameters of methods on the Phoenix-2014T \cite{camgoz2018neural}. Empty cells indicate that the information is not reported in the original papers.  \textit{VLP}: Visual-Language Pretraining, \textit{Light-T}: Lightweight translation model, \textit{CiCO}: Cross-Ingual COntrastive learning \cite{cheng2023cico}.\\
    \textsuperscript{†} A temporal module (Temporal Block + Temporal Encoder) follows ResNet18 in all cases. These temporal components differ across methods and are detailed in the sub-rows.
    \textsuperscript{*} These details are not reported explicitly in SignCL, but the authors state that all training settings follow those used in GFSLT. }
     \label{tab:parameters_sota} 
\end{table*}

\subsection{GFSLT-VLP}
\textbf{GFSLT-VLP} \cite{zhou2023gloss} introduces a two-stage SLT method with a novel visual-language pretraining strategy. In the first (pretraining) stage, sign language videos and their corresponding spoken language sentences are projected into a joint semantic space using contrastive learning. For the visual encoder, ResNet18 \cite{he2016deep} (2D-CNN) is followed by a temporal block (Conv1D-BN-Relu-MaxPooling) and 3-layers of the mBART \cite{liu2020multilingual} encoder. Textual representations are obtained with 12-layers of the mBART \cite{liu2020multilingual} encoder. These visual and textual embeddings are projected via linear layers into a shared embedding space, where contrastive learning maximizes similarity between paired inputs and minimizes it for mismatched pairs. Simultaneously, a masked self-supervised learning strategy is utilized to pretrain a Text Decoder (3-layers of the mBART decoder), aiming to capture the syntactic and semantic knowledge of spoken language sentences. In the second stage, an encoder-decoder model is fine-tuned for the SLT task, initialized with the pretrained visual encoder and text decoder.

\subsection{SignCL}
\textbf{SignCL} \cite{ye2024improving} identifies a representation density problem as visually similar but semantically distinct signs tend to be mapped to nearby locations in the feature space. To address this, they incorporate a contrastive learning loss between adjacent frames, encouraging the separation of different signs and improving representation quality. They assume that if two frames are close enough, they are considered to belong to the same sign, otherwise different signs. 

\subsection{Sign2GPT}
\textbf{Sign2GPT} \cite{wong2024signgpt} utilised pseudo-gloss pretraining, where glosses are generated from spoken sentences. During pretraining, visual features are aligned to pseudo-glosses instead of full sentences. For the visual encoder, Dino-V2 Vision Transformer is followed by a 4-layers transformer. In the second stage, XGLM, a multilingual Generative Pretrained Transformer (GPT) model is fine-tuned with LoRA (Low-Rank Adapters), a lightweight adaptation technique.

\subsection{FLa-LLM}
While the aforementioned approaches utilized contrastive learning to align video and text features in their pretraining stage, \textbf{FLa-LLM} \cite{chen2024factorized} employs a lightweight translation (Light-T) model while pre-training the visual encoder. This Light-T corresponds to the second stage of GFSLT, where a 3-layer mBART decoder is used to generate translations from video representations. In the second stage, the pretrained visual encoder is frozen, and a larger MBart (12 layers) is employed. Compared to GFSLT, this model differs in several architectural aspects -- such as applying frame subsampling instead of max pooling and reducing embedding dimensionality (see Table~\ref{tab:parameters_sota}).

\subsection{C\textsuperscript{2}RL}
\textbf{C\textsuperscript{2}RL} \cite{chen2024c} introduces multi-task pretraining framework by combining visual-text alignment and a lightweight translation objective. Similar to GFSLT, it aligns sign language videos and spoken language sentences in a shared semantic space, but further incorporates a Cross-Ingual COntrastive learning (CiCO) loss \cite{cheng2023cico}, referred to as Implicit Content Learning (ICL). Additionally, the Explicit Context Learning (ECL) focuses on translating video inputs to spoken language. 

\section{Evaluation Methodology}
In this section, we describe how the main contributions of each model were incorporated into our unified codebase during the reproduction process. Rather than re-implementing every detail, we focused on the core ideas proposed by each method under a standardized training setup.

\begin{itemize}
    \item \textbf{Visual-language pretraining: } Motivated by \cite{zhou2023gloss}, we adopt a two-stage SLT approach, combining contrastive visual-language pretraining and encoder-decoder fine-tuning.  In our setup, we retain its visual and textual encoder structure and pretraining objectives. The architecture of our baseline is illustrated in Figure~\ref{fig:main}.

\begin{figure*}[htbp]
    \centering
    \includegraphics[width=\linewidth]{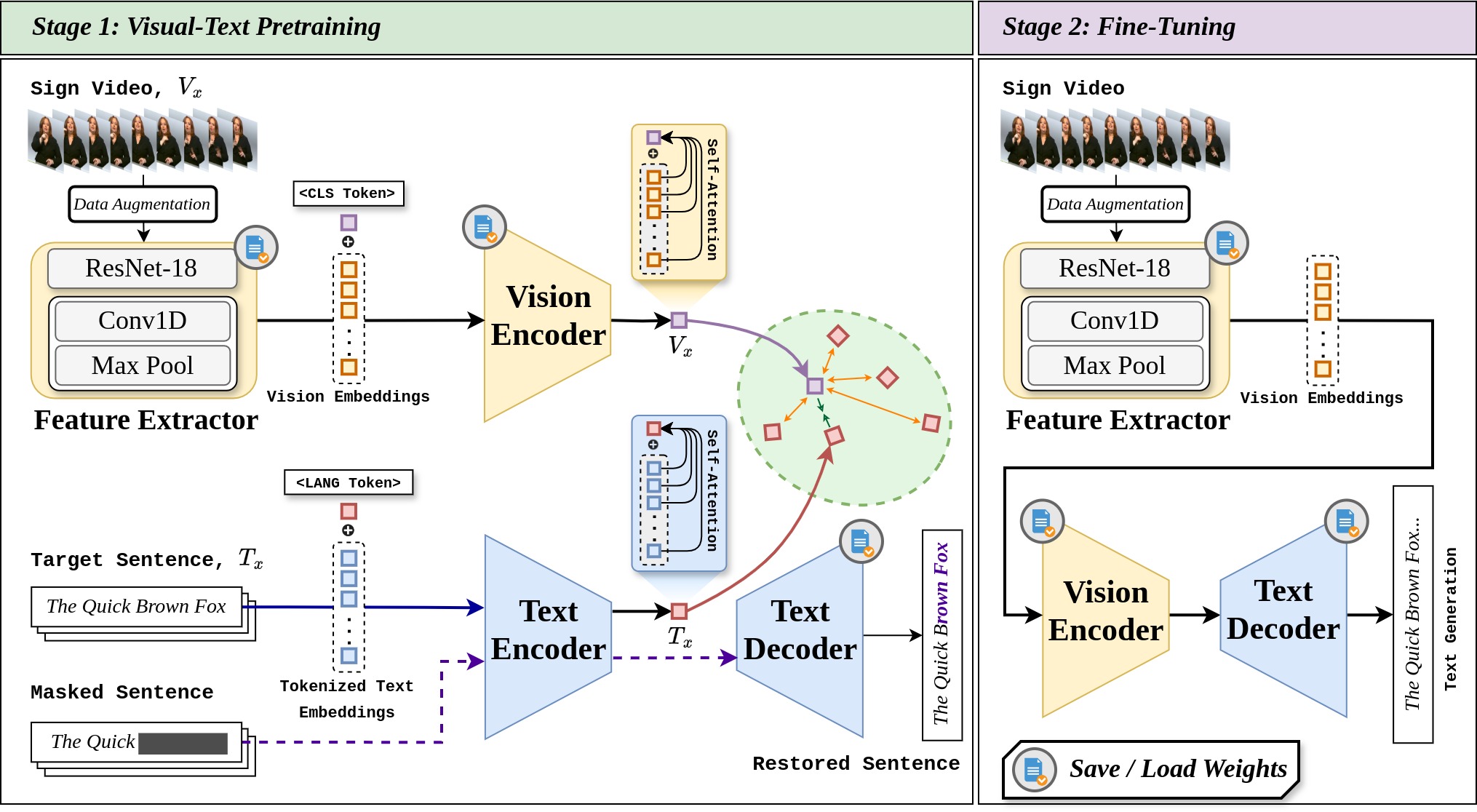}        
    \caption{Overview of our two-stage architecture. \textbf{\textit{Pretraining stage:}} Video frames are encoded using a ResNet-18 backbone, followed by temporal modeling with 1D convolutions and max pooling. A [CLS] token is appended and passed through a Transformer encoder to obtain a global video representation. Text is encoded similarly and summarized into a [LANG] token. Contrastive learning is then employed to match CLS-LANG token pairs. Meanwhile, a masked sentence modeling objective is used to train the text decoder. \textbf{\textit{Fine-tuning stage:}} The pretrained visual encoder, text encoder, and decoder are loaded from pretraining and fine-tuned for sign translation using standard sequence generation.}
  \label{fig:main}
\end{figure*}

    \item \textbf{Frame-level contrastive learning } is integrated into our baseline network to learn more discriminative feature representation in a self-supervised manner. We adopt the loss formulation from the official codebase provided by the authors \cite{ye2024improving}. The loss is added to both the pretraining and fine-tuning stages.
    
    \item \textbf{Pseudo-gloss pretraining: } We first create gloss-like intermediate representations, pseudo-glosses from spoken language sentences utilizing the official codebase \cite{wong2024signgpt}. Then, we follow the pseudo-gloss pretraining strategy instead of the base CLIP-style loss. However,  we keep our visual backbone consistent, such as Resnet18, Conv1D-BN-Relu-MaxPooling, and a transformer encoder. 
    
    \item \textbf{Lightweight translation, Light-T: } To implement the key idea of factorized learning \cite{chen2024factorized} within our unified framework, we first follow the translation task baseline setup, which aligns with the lightweight translation (Light-T) objective, employing a 3-layer mBART decoder to generate translations from video features. In the second stage, we freeze the pretrained visual encoder and replace the decoder with a larger, 12-layer mBART to enhance language modeling capacity. As opposed to the original paper, which uses a lower embedding size in the visual encoder, applies a 25\% frame sampling strategy, and trains with a larger batch size, we retain our standardized settings.
    
    \item \textbf{Cross-lingual Contrastive learning (CiCO) loss in pretraining:} As the first contribution of C\textsuperscript{2}RL \cite{chen2024c}, the Implicit Content Learning (ICL), we replace the original CLIP-style contrastive loss with CiCO loss \cite{cheng2023cico} in pretraining. In the second stage, we perform standard SLT fine-tuning using the pretrained visual encoder and a 3-layer mBART decoder.
    \item \textbf{Multi-task pretraining: } We refer to the core contribution of C\textsuperscript{2}RL, namely the combination of Implicit Content Learning (ICL) and Explicit Content Learning (ECL), as a multi-task pretraining strategy. Specifically, we jointly apply the CiCO loss for visual-text alignment and a lightweight translation objective using a 3-layer mBART decoder. For the second stage, we load only Resnet18, Conv1D-BN-Relu-MaxPooling as the visual encoder.  We fine-tune a 12-layer mBART decoder while keeping the pretrained visual encoder frozen.

\end{itemize}

%% file: sections/4_experiments.tex
\section{Results}

This section presents the results of evaluating several state-of-the-art gloss-free sign language translation (SLT) approaches. The goal of this evaluation is to provide a critical comparison of existing methods. We trained and tested all models under identical conditions, ensuring a fair and consistent comparison to measure the effectiveness of their design decisions.

\subsection{Evaluation Metrics} 
We use BLEU \cite{papineni2002bleu} and ROUGE \cite{lin2004rouge} metrics to evaluate the performance of our SLT approach. BLEU (Bilingual Evaluation Understudy) measures the overlap between generated translations and reference sentences by computing precision over n-grams (consecutive word sequences). Particularly, BLEU-4 becomes the most commonly used variant in SLT. ROUGE-L (Recall-Oriented Understudy for Gisting Evaluation-Longest Common Subsequence), on the other hand, evaluates the quality of a generated sentence by measuring the length of the longest common subsequence between the candidate and reference texts. Unlike n-gram based metrics, ROUGE-L captures sentence-level structure, making it suitable for evaluating translations with flexible word order.
%while BLEURT aims to achieve human-quality scoring. 

\subsection{Implementation Details} 
\label{section:implementation_details}
We train all models on a single NVIDIA A100 GPU with a batch size of 8. We begin our experiments on the Phoenix-2014T dataset, initially adopting hyperparameters from the GFSLT-VLP implementation \cite{zhou2023gloss}. In pretraining, the original GFSLT used a batch size of 16 with a learning rate of $10^{-2}$. Since we reduce the batch size to 8 due to hardware constraints, we proportionally lower the learning rate to $5 \times 10^{-3}$ during pretraining. For the translation stage, where both the original and our setup use a batch size of 8, we maintain the original learning rate of $10^{-2}$. This configuration serves as our baseline, and we strive to maintain consistency across all methods and datasets.

\begin{figure}[h!]
    \centering
    \begin{subfigure}[b]{0.33\textwidth}
        \includegraphics[width=\textwidth]{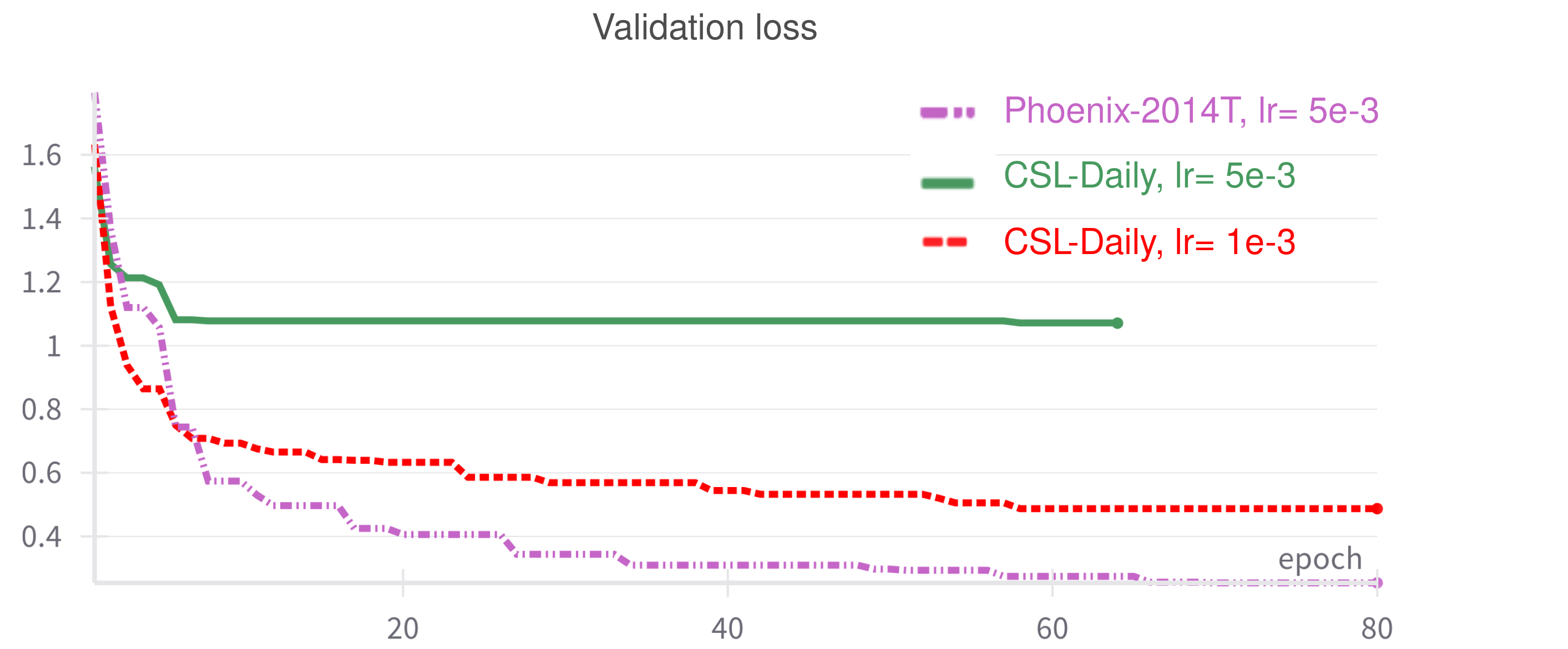}
        \caption{Pretraining stage}
        \label{fig:p}
    \end{subfigure}
    \hfill
    \begin{subfigure}[b]{0.66\textwidth}
        \includegraphics[width=\textwidth]{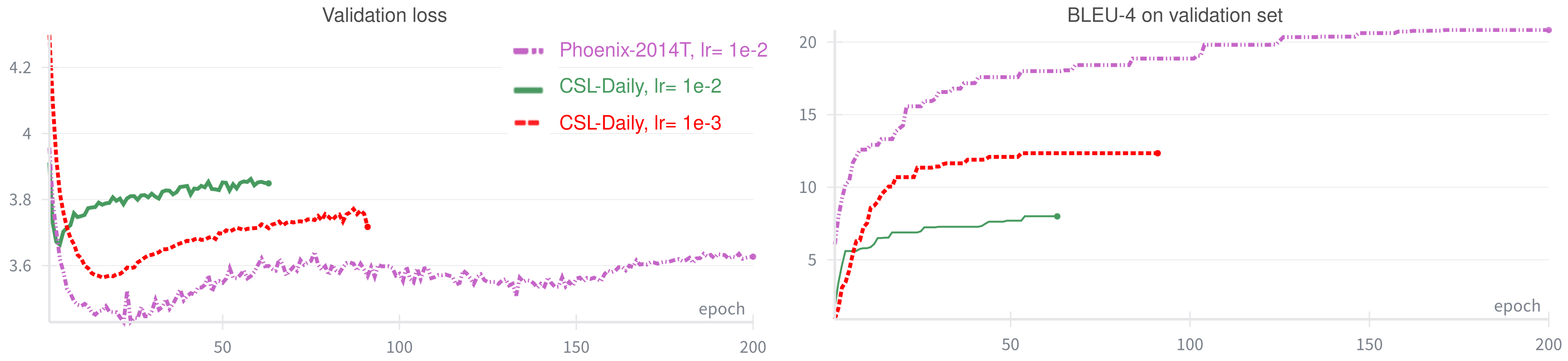}
        \caption{Translation stage}
        \label{fig:t}
    \end{subfigure}
    \caption{Validation loss and BLEU-4 score graphs for the GFSLT-VLP baseline \cite{zhou2023gloss}.  For CSL-Daily, validation loss remained stable or began to rise early in training when using the same learning rate as for Phoenix-2014T, suggesting that the learning rate was too high.}
    \label{fig:lr}
\end{figure}

However, \textit{exact consistency was not achievable across different datasets} as seen in Table \ref{tab:parameters}. While reproducing the baseline results on CSL-Daily, we observed that validation loss began to rise early in training (Figure \ref{fig:lr}), suggesting the \textbf{learning rate} was too high. We empirically reduced the learning rate to $10^{-3}$ for both pretraining and translation stages on CSL-Daily. 
\begin{table*}[h]
	\centering
    \resizebox{\linewidth}{!}{
    \begin{tabular}{l|ccc|ccc}
        \hline   
        \textbf{} & \multicolumn{3}{c}{\textbf{Phoenix-2014T}} & \multicolumn{3}{c}{\textbf{CSL-Daily}}  \\
        \hline
         & \textbf{GFSLT, SignCL} & \textbf{FLa-LLM} & \textbf{C\textsuperscript{2}RL} & \textbf{GFSLT, SignCL} & \textbf{FLa-LLM} & \textbf{C\textsuperscript{2}RL}  \\
        \hline

        \rowcolor[HTML]{E2E2E2}
        \multicolumn{7}{l}{Pretraining stage}  \\ 
        number of epochs & 80 & 100 & 200 & 80 & 100 & 100  \\ 
        learning rate & $5 \times 10^{-3}$  & $10^{-2}$ 
        & $10^{-2}$
        & $5 \times 10^{-3}$, \underline{$10^{-3}$} & 
        $\underline{5 \times 10^{-3}}, 10^{-3}$ & $\underline{5 \times 10^{-3}}, 10^{-3}$  \\    

        \rowcolor[HTML]{E2E2E2}
        \multicolumn{7}{l}{Translation stage}  \\ 
        number of epochs & 200 & 75 & 80 & 80  & 75 & 80  \\
        learning rate & $10^{-2}$  
        & $10^{-2}$, \underline{$5\times 10^{-3}$}
        & $\underline{10^{-2}}, 5 \times 10^{-3}$  & $10^{-2}$, \underline{$10^{-3}$}& $5 \times 10^{-3}$ &  $\underline{5 \times 10^{-3}}, 10^{-3}$   \\ 

        \hline  
    \end{tabular}
    }
 	\caption{Hyperparameters of different datasets across methods. Underlined values indicate the best-performing learning rate.}
    \label{tab:parameters} 
\end{table*}

In addition, the \textbf{training time} varied significantly across datasets. We observed that models converged much earlier on CSL-Daily, with minimal improvement beyond 50 epochs. Therefore, we trained models on CSL-Daily up to 80 epochs in the second phase. We hypothesize that early convergence is because CSL-Daily is larger, has cleaner videos recorded in a lab setting, and includes some sentences that are repeated by different signers.

Another consideration is the variation in \textbf{sequence lengths}. Both Phoenix-2014T and CSL-Daily contain videos typically fewer than 300 frames. While GFSLT-VLP and SignCL restricted the maximum number of frames to 300 by random sampling, other methods did not report such details. We applied the same restriction to all methods to ensure a fair comparison.

All models are trained and evaluated three times with different random seeds (0, 42, and 100) to ensure robustness and reproducibility of results.

%---------------------------

\subsection{Experimental Comparison of Models}

We successfully reproduce GFSLT-VLP \cite{zhou2023gloss} results on Phoenix-2014T, with our BLEU-2, BLEU-3, and BLEU-4 scores closely aligning with the originally reported values (Table \ref{tab:phoenix}), and our reproduction achieves higher BLEU-1 (43.71 vs 46.44 ± 0.48) and ROUGE (42.49 vs 46.87 ± 0.58). For CSL-Daily, we use the adjusted learning rate and shortened training time as described in Section \ref{section:implementation_details}. However, our results on CSL-Daily are slightly below those reported in the original paper, as seen in Table \ref{tab:csl} (BLEU-4: 11.00 vs 10.41). 

   \begin{table*}
    \centering
    \resizebox{1\linewidth}{!}{
        \begin{tabular}{ lccccc }
        \hline  

        \textbf{Method}  & \textbf{BLEU1}& \textbf{BLEU2}& \textbf{BLEU3} & \textbf{BLEU4} &  \textbf{ROUGE-L}     \\ 
        
        \hline
        \rowcolor[HTML]{E2E2E2}
        \multicolumn{6}{l}{Reported}  \\ 
        GFSLT-VLP \cite{zhou2023gloss} & 43.71 & 33.18 & 26.11 & 21.44  & 42.49   \\
        Sign2GPT \cite{wong2024signgpt} &49.54 & 35.96 & 28.83 & 22.52  & 48.94\\
        SignCL \cite{ye2024improving}  & 49.76 & 36.85 & 29.97 & 22.74   & 49.04   \\
        FLa-LLM \cite{chen2024factorized} &  46.29 & 35.33 & 28.03 & 23.09  & 45.27 \\
        C\textsuperscript{2}RL (ICL only) \cite{chen2024c}  & - & - & - & 24.83   \\
        C\textsuperscript{2}RL (ICL + ECL) \cite{chen2024c}  & 52.81 & 40.20 & 32.20 & 26.75  & 50.96 \\ \hline

        \rowcolor[HTML]{E2E2E2}
        \multicolumn{6}{l}{Reproduced}  \\ 
      Visual-language pretraining \cite{zhou2023gloss} & 46.44 ± 0.48 & 33.96 ± 0.40 & 26.69 ± 0.32 & 21.97 ± 0.27 & 46.87 ± 0.58  \\
    Contrastive learning on frames \cite{ye2024improving} & 46.62 ± 0.28 & 34.41 ± 0.15 & 27.06 ± 0.22 & 22.35 ± 0.28 & 47.30 ± 0.26 \\
    Lightweight translation (Light-T) in pretraining \cite{chen2024factorized} &  45.29 ± 0.26 & 32.08 ± 0.29 & 24.53 ± 0.33 & 19.84 ± 0.28 & 44.43 ± 0.15 \\
    Utilize CICO loss in pretraining \cite{chen2024c} & 47.30 ± 0.47 & \textbf{35.05 ± 0.51} & \textbf{27.72 ± 0.59} & \textbf{22.95 ± 0.63} & \textbf{48.20 ± 0.37} \\
    CICO loss and Light-T task in pretraining \cite{chen2024c} & \textbf{47.61 ± 0.35} & 34.41 ± 0.42 & 26.48 ± 0.43 & 21.41 ± 0.43 & 46.84 ± 0.72  \\   \hline   
    \end{tabular}
    }
    \caption{Reported and reproduced results on the Phoenix-2014T. }
    \label{tab:phoenix} 
\end{table*}

\begin{table*}
    \centering
    \resizebox{1\linewidth}{!}{
        \begin{tabular}{ lccccc }
        \hline  

        \textbf{Method}  & \textbf{BLEU1}& \textbf{BLEU2}& \textbf{BLEU3} & \textbf{BLEU4} &  \textbf{ROUGE-L}   \\ 
        
        \hline
        \rowcolor[HTML]{E2E2E2}
        \multicolumn{6}{l}{Reported}  \\ 
        GFSLT-VLP \cite{zhou2023gloss} &  39.37 & 24.93 & 16.26 & 11.00 & 36.44  \\
        Sign2GPT \cite{wong2024signgpt} & 41.75 & 28.73 & 20.60 & 15.40 & 42.36 \\
        SignCL \cite{ye2024improving}  & 47.47 & 32.53 & 22.62 & 16.16 & 48.92 \\
        FLa-LLM \cite{chen2024factorized} & 37.13 & 25.12 & 18.38 & 14.20  & 37.25 \\
        C\textsuperscript{2}RL (ICL + ECL) \cite{chen2024c} & 49.32 & 36.28 & 27.54 & 21.61 &  48.21 \\
        \hline

        \rowcolor[HTML]{E2E2E2}
        \multicolumn{6}{l}{Reproduced}  \\ 
        Visual-language pretraining \cite{zhou2023gloss} & 36.78 ± 0.67 & 23.23 ± 0.50 & 15.25 ± 0.34 & 10.41 ± 0.17 & 35.60 ± 0.65  \\
        Contrastive learning on frames \cite{ye2024improving} & 37.31 ± 0.26 & 23.97 ± 0.26 & 16.11 ± 0.21 & 11.22 ± 0.18 & 36.42 ± 0.12  \\ 
        Lightweight translation (Light-T) in pretraining \cite{chen2024factorized}  & 37.41 ± 1.33 & 23.30 ± 1.07 & 15.22 ± 0.70 & 10.26 ± 0.40 & 36.69 ± 1.11  \\ 
        Utilize CICO loss in pretraining \cite{chen2024c} & \textbf{39.98 ± 0.18} & \textbf{26.06 ± 0.13} & \textbf{17.45 ± 0.06} & \textbf{12.03 ± 0.06} & \textbf{38.42 ± 0.17}  \\ 
        CICO loss and Light-T task in pretraining \cite{chen2024c} & 39.59 ± 0.10 & 25.39 ± 0.19 & 16.77 ± 0.25 & 11.51 ± 0.27 & 37.36 ± 0.34 \\ \hline
    \end{tabular}
    }
    \caption{Reported and reroduced results on the CSL-Daily. }
    \label{tab:csl} 
\end{table*}

%We use GFSLT-VLP as a baseline and re-implement the key components of each method under a consistent experimental setting. All models are trained and evaluated three times with different random seeds (0, 42, and 100). Table~\ref{tab:phoenix} presents the average performance along with standard deviation across these runs.

Next, we added the contrastive learning loss for adjacent frames from SignCL \cite{ye2024improving}. Our reproduced results showed improvements over the GFSLT-VLP baseline, but did not reach the reported results in the original paper. This issue has also been noted by other researchers. We believe the possible reason for this discrepancy is the use of a smaller batch size (8 rather than 128 as reported), which may have limited the effectiveness of the contrastive loss. Nevertheless, additional contrastive learning yielded a lower standard deviation across seeds, indicating better training stability.

We then attempted to implement pseudo-gloss pretraining into our architecture. However, this configuration yielded significantly lower performance compared to the original paper and was therefore excluded from our results table. To validate the reproducibility of Sign2GPT \cite{wong2024signgpt}, we also ran their original codebase, which includes components such as DINOv2 \cite{oquab2023dinov2}, LoRA (Low-Rank Adapters), and the XGLM language model \cite{lin2022-shot}. Using their official setup, we were able to reproduce results that align closely with those reported in the original paper. This suggests that the performance of Sign2GPT \cite{wong2024signgpt} depends on its specific architectural and training choices.

\begin{figure}[t]
    \centering
    \begin{subfigure}[b]{0.48\textwidth}
        \includegraphics[width=\textwidth]{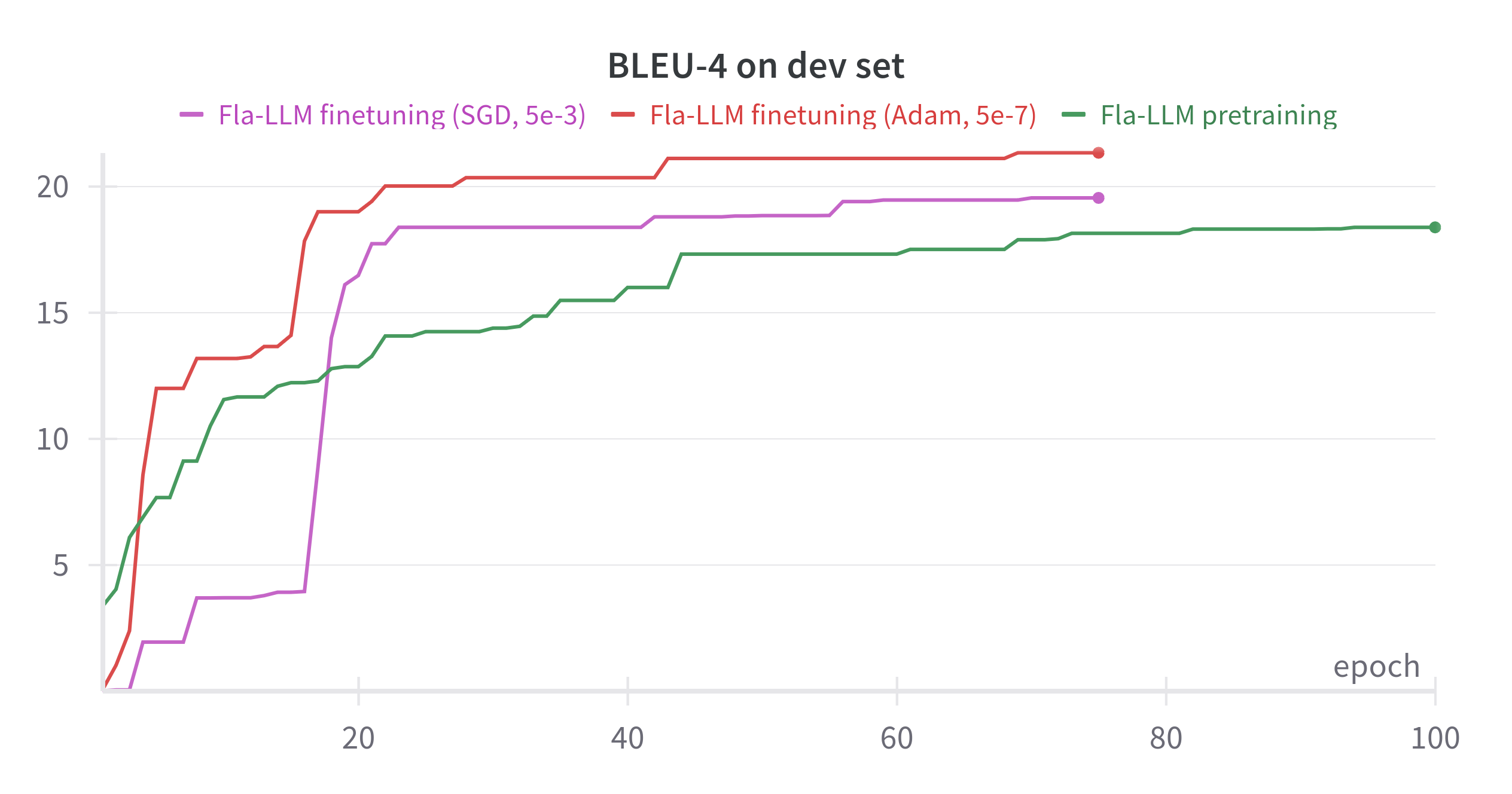}
        \caption{FLa-LLM: Light translation during pretraining}
        \label{fig:fla}
    \end{subfigure}
    \hfill
    \begin{subfigure}[b]{0.48\textwidth}
        \includegraphics[width=\textwidth]{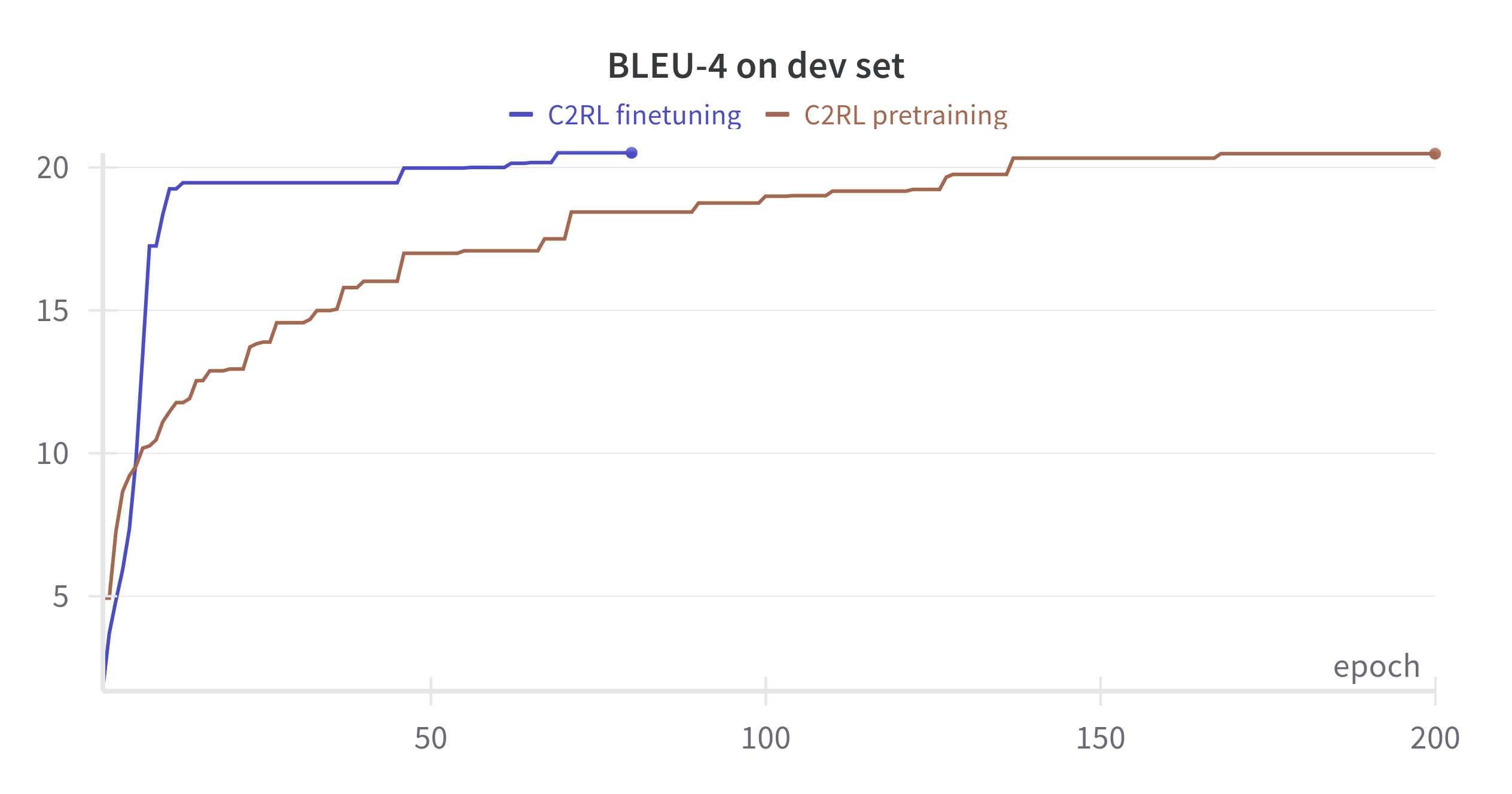}
        \caption{C\textsuperscript{2}RL: CiCO loss and light translation during pretraining}
        \label{fig:c2rl}
    \end{subfigure}
    \caption{BLEU-4 progression on the Phoenix-2014T dev set for FLa-LLM \cite{chen2024factorized} and C\textsuperscript{2}RL \cite{chen2024c} during training. Each model is first pretrained and then fine-tuned starting from the best pretrained checkpoint. Fine-tuning exhibits rapid improvement, but then saturates. This behavior causes the gap between our reproduced results and those reported in the original papers.}
    \label{fig:fla_c2rl}
\end{figure}

Our next experiment explores the use of lightweight-translation during pretraining stage, which is the key contribution proposed in FLa-LLM \cite{chen2024factorized}. To this end, we used the translation task setup from the GFSLT-VLP baseline and obtained a BLEU-4 score of 19.50 (Table \ref{tab:flallm_ablation}). This score is expected, as it reflects the performance of the SLT model without any visual-language pretraining. In the subsequent fine-tuning stage, although learning progressed quickly at the beginning, BLEU scores
plateaued shortly after (see Fig.~\ref{fig:fla_c2rl}). The final BLEU-4 score of 19.84 ± 0.28 remained below the baseline, likely due to early overfitting caused by the increased model capacity. These results were obtained using the same optimizer, SGD, as in our earlier experiments. To better understand the performance gap, we conducted an ablation study using the Adam optimizer, as employed in the original paper, on the Phoenix-2014T dataset. A comparison of BLEU-4 scores across different learning rates using Adam is provided in Table~\ref{tab:flallm_ablation}. While Adam leads to better performance than SGD, it was still insufficient to close the gap with the reported results. Several other factors may have contributed to this discrepancy. For instance, FLa-LLM employs a lower embedding size in the visual encoder, applies a 25\% frame sampling strategy, and uses a larger batch size. The authors' own ablation studies show that both of these choices positively affect performance.

\begin{table*}
    \centering
    \resizebox{1\linewidth}{!}{
        \begin{tabular}{ lccccccc }
        \hline  

         \textbf{Stage}& \textbf{Optimizer}& \textbf{Learning rate} & \textbf{BLEU1}& \textbf{BLEU2}& \textbf{BLEU3} & \textbf{BLEU4} &  \textbf{ROUGE-L}   \\ 
        
        \hline
         Pretraining & SGD & $10^{-2}$ & 44.63 & 31.72 & 24.15 & 19.50  & 43.69 \\ \hline
        Fine-tuning & SGD & $5 \times 10^{-3}$  & 45.29 ± 0.26 & 32.08 ± 0.29 & 24.53 ± 0.33 & 19.84 ± 0.28 & 44.43 ± 0.15  \\ 
        & Adam & $5 \times 10^{-6}$   & 44.44 & 32.13 & 24.89 & 20.24  & 44.47 \\
        & Adam & LLM: $5 \times 10^{-6}$, LLM Adapter: $5 \times 10^{-4}$   & x & x & x & x & x \\
        & Adam & $5 \times 10^{-7}$  & 46.55 & 33.76 & 26.16 & 21.09 & 44.86  \\
     & Adam & LLM: $5 \times 10^{-7}$, LLM Adapter: $5 \times 10^{-5}$ & 45.37 & 32.75 & 25.14  & 20.23 & 43.51 \\ \hline
    \end{tabular}
    }
    \caption{Performance of FLa-LLM \cite{chen2024factorized} on Phoenix-2014T using different learning rate variations with Adam optimizer. The experiment marked with `x' was aborted early due to unstable training caused by a high learning rate.}
    \label{tab:flallm_ablation} 
\end{table*}

In our first experiment with the C\textsuperscript{2}RL setup, we replaced the original CLIP-based visual-text alignment loss with the CiCO loss \cite{cheng2023cico}, which is designed to better capture semantic relationships across sign and spoken modalities. This substitution resulted in improved translation performance, achieving BLEU-4 scores of 22.95 (vs. 21.97) on Phoenix-2014T and 12.03 (vs. 10.41) on CSL-Daily.  It suggests that CiCo provides a more effective supervision signal for aligning video and text representations.

Next, we combined CiCo-based alignment with lightweight translation during pretraining. Compared to the single-task setup used in FLa-LLM—where only lightweight translation was performed—this multi-task configuration yielded better initial performance. However, as with FLa-LLM, further improvements could not be achieved in the second stage. Although learning initially progressed rapidly, BLEU scores quickly plateaued. This behavior may point to overfitting or limited transferability beyond the pretraining stage. 

We therefore believe that the design choices in the original paper, such as applying dropout in the second stage, and adopting a larger batch size, may have contributed to mitigating overfitting and achieving higher BLEU scores. In contrast, we deliberately refrained from making such architectural modifications introduced in their work to ensure consistency across our own experiments and to isolate the effect of the key contribution, the multi-task pretraining strategy.

% --------------------------------------------------
\subsection{Metric Inconsistency in Evaluation}

\textbf{Dataset-Specific Conventions: } 
We observed that metric calculations differ between datasets. For instance, although Phoenix-2014T does not contain punctuation in its original sentences, many works append a dot with a space (` .') to each sentence, artificially inflating BLEU scores. On the other hand, in CSL-Daily, character-based BLEU is often reported, which produces again higher scores than word-based evaluation.  These implementation-specific conventions are rarely stated in papers and need to be inferred from qualitative results or the available implementations. To ensure a fair comparison with prior works, we follow the same conventions when reporting results on each dataset. 

\textbf{Library-Specific Variability: } We also found that some metric scores may not be consistent across libraries, particularly between popular toolkits such as \texttt{nlg-eval} and \texttt{sacreBLEU}. This discrepancy arises in CSL-Daily, where character-based segmentation is used. For example, the same baseline model \cite{zhou2023gloss} can yield a BLEU-4 score of 10.41 using \texttt{nlg-eval} (v2.4.1) and 11.66 using \texttt{sacreBLEU} \footnote{The exact signature for the sacreBLEU is:  \texttt{nrefs:1|case:mixed|eff:no|tok:13a|smooth:exp|version:2.2.0.} \texttt{tok:zh} also produced the same results.}. In contrast, Phoenix-2014T does not exhibit such variation in our experiments. For all aforementioned experiments, we used \texttt{nlg-eval}  to ensure a fair comparison.

%% file: sections/5_conclusion.tex
%------------------------------------------------------------------------
\section{Conclusion}

This paper presented a systematic comparison of recent gloss-free sign language translation models, which differ in their pretraining strategies. These strategies include vision-language pretraining, contrastive learning on adjacent frames, pseudo-gloss supervision, lightweight translation modules during pretraining, and the combination of visual-language alignment with lightweight translation.

By re-implementing all models within a unified, modular codebase and evaluating them under identical training conditions, we isolate the contribution of each pretraining approach from other factors such as data preprocessing and optimization schedules. Our experiments show that performance gaps reported in the literature often narrow under controlled conditions. Still, we observe that some pretraining designs, such as employing cross-lingual contrastive learning to align sign videos with textual representations, yield better results.

We release our codebase to promote transparency, reproducibility, and future experimentation in sign language translation research. This work provides a standardized and modular framework that can serve as a helpful starting point for further studies. Future directions include scaling the models and evaluation to larger sign language datasets to improve generalization across different signers and domains.